\documentclass[letterpaper, 10 pt, conference]{ieeeconf}  

\IEEEoverridecommandlockouts                              

\usepackage{cite}
\usepackage{amsmath,amssymb,amsfonts}
\usepackage[ruled,vlined]{algorithm2e}
\usepackage{graphicx}
\usepackage{textcomp}
\usepackage{epsfig} 
\usepackage{newtxtext,newtxmath}
\usepackage{blindtext}
\usepackage[hidelinks]{hyperref}

\usepackage{soul,xcolor}
\usepackage{multirow}
\usepackage{colortbl}
\usepackage{tabularray}
\usepackage{hhline}
\usepackage{bm}
\usepackage{makecell}
\usepackage{booktabs}
\usepackage{comment}
\usepackage{stfloats}
\usepackage[inkscapelatex=false]{svg}
\usepackage{tabularx}
\usepackage{booktabs}
\usepackage{threeparttable}

\def\BibTeX{{\rm B\kern-.05em{\sc i\kern-.025em b}\kern-.08em
    T\kern-.1667em\lower.7ex\hbox{E}\kern-.125emX}}

\title{\LARGE \bf
Scalable, Intuitive, Mixed Reality–Based Semi-Autonomous Robotic Telemanipulation for the Execution of a Fine-Grained Bimanual Task}

\begin{document}

\author{Shifei Duan, Zhe Wang, Francesco De Pace, and Minas Liarokapis
\thanks{Shifei Duan and Zhe Wang are with the New Dexterity research group, The University of Auckland, New Zealand
        {\tt\footnotesize  \{sdua078, zwan341\}@aucklanduni.ac.nz}}%
\thanks{Francesco De Pace is with the Competence Industry Manufacturing 4.0 (CIM4.0), Turin, Italy {\tt\footnotesize francesco.de\_pace@cim40.com}}%
\thanks{Minas Liarokapis is with the New Dexterity research group, The University of Auckland, New Zealand and the National Technical University of Athens, Greece
        {\tt\footnotesize liarokapis@mail.ntua.gr}}}

\maketitle
\thispagestyle{empty}
\pagestyle{empty}

\begin{abstract}

Nowadays, robotic telemanipulation enables humans to control robots so as to execute tasks in a manner similar to how they perform them, and serves as an essential approach for collecting on-robot data for robot learning by human demonstrations. The quality of collected demonstrations directly impacts the performance of the learned policy. However, existing telemanipulation systems rely heavily on human operation, particularly for large-scale data collection. Collecting dozens or hundreds of demonstrations is a demanding process, as operators tend to fatigue quickly, especially during the execution of dexterous tasks that require sustained concentration. Recently, semi-autonomous robotic telemanipulation has emerged as a promising direction that reduces operator workload while preserving human-in-the-loop control for task execution. We therefore introduce a scalable, intuitive, semi-autonomous robotic telemanipulation system that leverages a mixed reality (MR) headset for on-robot data collection. The system autonomously executes the majority of the task trajectories and only requires the operator to make fine adjustments at key stages. We validate the effectiveness of our system by collecting data and training imitation learning policies on the execution of a high-precision, fine-grained bimanual peg-in-hole task. With lower data collection effort, our system obtains high-quality demonstrations that can be used to train fully autonomous task execution policies.
\end{abstract}

\section{Introduction}


Recent years have witnessed significant progress in learning-based robotic manipulation, where the availability of large-scale real-robot datasets has significantly boosted robot deployment across a wide range of applications, from daily tasks to industrial scenarios\cite{brohan2022rt}. Among the various methods for collecting data, telemanipulation has proven particularly valuable for imitation learning, providing high-quality demonstrations characterized by natural and smooth trajectories that are essential for policy generalization\cite{iyer2024open}. Various telemanipulation systems have been investigated, including virtual reality (VR) devices\cite{lin2025learning}, cameras\cite{qin2023anyteleop}, and wearable gloves\cite{caeiro2021systematic}. VR devices in particular have been widely adopted, offering immersive three-dimensional workspace representations that enhance spatial awareness and task execution accuracy\cite{hetrick2020comparing, park2021study}. More recently, the emergence of advanced Mixed Reality (MR) platforms, particularly the Apple Vision Pro, has renewed interest in telemanipulation research, as its high-resolution displays, rich sensor suite, and onboard processing capabilities open new avenues for immersive robotic operation\cite{ding2025bunny, cheng2024open}.

Despite its advantages, telemanipulation presents notable challenges, particularly for the execution of fine-grained manipulation tasks that require subtle adjustments that are difficult for operators to perceive and execute accurately. Furthermore, large-scale data collection requires operators to repeat the same task numerous times so as to acquire sufficient demonstrations, making it increasingly difficult to maintain concentration throughout the entire process.
Fully autonomous systems, although effective for repetitive tasks in structured environments, often struggle in unstructured or unfamiliar scenarios demanding high dexterity. While vision-based autonomous systems have made considerable progress in addressing these challenges, ensuring reliable execution without any human supervision remains difficult, as even minor errors can lead to task failure.
\begin{figure}[t]
    \centering
    \includegraphics[width=0.65\columnwidth]{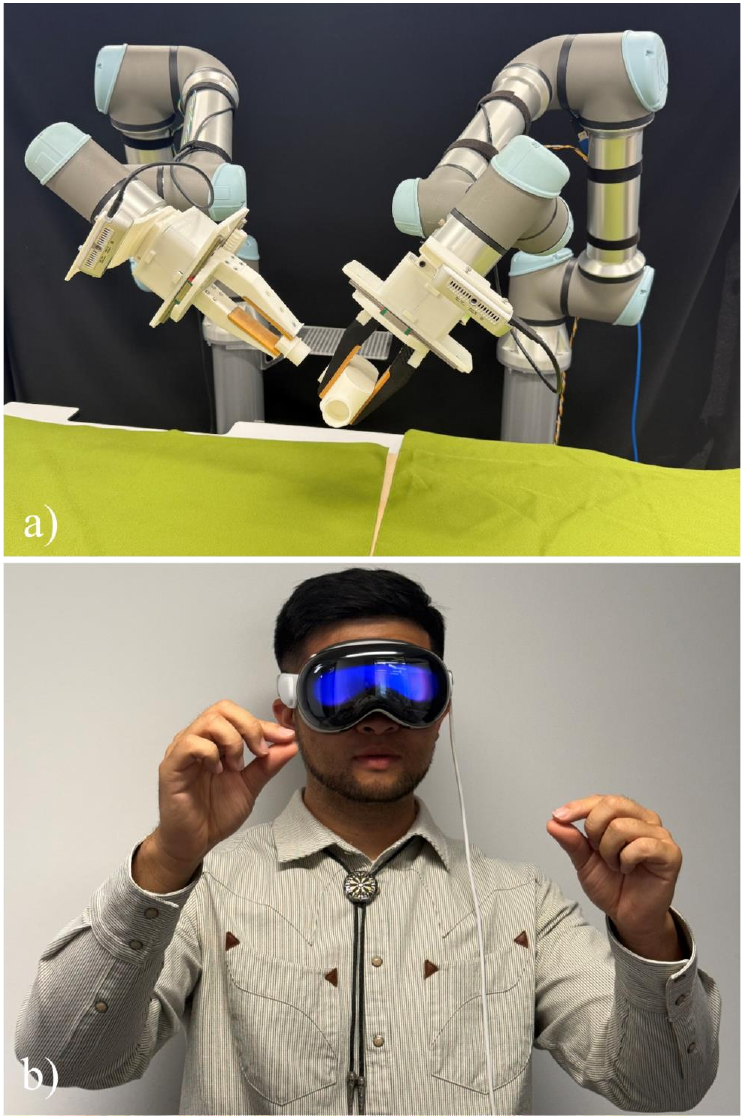}
    \caption{Subfig. a) The bimanual robotic system executing the fine-grained manipulation task. Subfig. b) A user wearing the MR headset during the collection of on-robot demonstrations, using the hands for robot control.}
    \label{fig:setup}
\end{figure}
To address these complementary limitations, semi-autonomous systems have emerged as a promising approach\cite{alharthi2024towards,min2025semi}, in which human-in-the-loop control allows operators to leverage their natural perception and decision-making capabilities to detect and correct errors in real time, while the robot autonomously handles the remaining execution\cite{duan2024semi,duan2026comparing}. Although prior work has demonstrated the feasibility of such systems, their application to fine-grained bimanual tasks remains an open challenge. Specifically, bimanual tasks offer much lower error tolerance, as any misalignment in one arm is often accumulated by the required synchronization with the other, leading to rapid task failure.

Therefore, this paper introduces a Mixed Reality-based semi-autonomous robotic telemanipulation system designed for fine-grained manipulation tasks. As illustrated in Fig.\ref{fig:setup}, by capturing the user's hand gestures and movements through the Apple Vision Pro (AVP), our system translates them into precise robotic control commands in real time. The proposed system addresses existing limitations by integrating human-in-the-loop control with autonomous robotic execution, reducing operator workload while enabling efficient collection of large-scale high-accuracy demonstrations. The collected demonstrations are used to train an Action Chunking with Transformers (ACT) policy, validating the quality of the collected data and demonstrating the scalability of the proposed pipeline.

\section{Related work}
\label{sec:soa}

\textbf{Telemanipulation.} Recent advances in learning-based robotic manipulation have driven increasing demand for large-scale data collection, leading to the development of diverse telemanipulation systems. Among these, virtual reality-based interfaces have gained particular attention for enabling immersive and intuitive remote control of robotic manipulators. Various VR-based telemanipulation systems have been studied for collecting human demonstrations for robot learning\cite{zhang2018deep,delpreto2020helping,pan2021augmented,iyer2024open}. Building on this, mixed reality-based telemanipulation further enhances the operator experience by overlaying virtual information onto the real environment, enabling more intuitive control and richer situational awareness\cite{penco2024mixed}. Several studies have leveraged the AVP headset for dexterous telemanipulation, achieving improved data collection efficiency by utilizing its high-quality display and built-in hand-tracking capabilities\cite{cheng2024open,ding2025bunny}. Moreover, the AVP has been shown to outperform traditional VR headsets and other teleoperation interfaces in terms of usability and operator performance\cite{duan2026mr}. Nevertheless, these systems often require considerable telemanipulation experience from the operator, and the tasks demonstrated tend to be relatively simple, failing to meet the precision requirements of industrial applications. In order to solve more complex tasks while maintaining usability, semi-autonomous telemanipulation has emerged as a direction. Previous work has demonstrated its effectiveness in complex assembly tasks, with standardized user studies including the System Usability Scale (SUS) and NASA Task Load Index (NASA-TLX) showing that semi-autonomous telemanipulation offers notable improvements over pure telemanipulation in terms of physical demand, mental workload, and overall user experience\cite{duan2024semi,duan2026comparing}. Consequently, rather than pursuing dexterous hand-based systems focused on daily tasks, our work targets industrial scenarios where parallel grippers are sufficient for most operations\cite{li2025comparative}. Compared to dexterous hands, parallel grippers reduce the degrees of freedom involved, minimizing accumulated errors and yielding simpler and more predictable grasping behavior.

\textbf{Data Collection.} Collecting real-robot demonstrations has been shown to be more effective than simulation-based approaches\cite{khazatsky2024droid,brohan2022rt,zitkovich2023rt,siekmann2021blind}, primarily due to the large sim-to-real gap arising from complex physical contacts between the robot end-effector (EE) and objects. A classic approach involves exoskeleton devices or mirroring arms for joint-space mapping\cite{zhao2023learning,fang2023low,fu2024mobile,wu2024gello}, where the follower arm directly replicates the motion of the leader arm, enabling precise control for tasks such as battery insertion and velcro fastening\cite{zhao2023learning}. However, these systems rely entirely on manual operation, requiring operators to control every stage of the task, which leads to fatigue and limits the scalability of data collection, motivating the requirement for more efficient and accessible data collection approaches.

\textbf{Imitation Learning.} Imitation learning enables robots to learn manipulation skills directly from expert demonstrations. Early approaches based on behavioral cloning cast imitation as supervised learning from observations to actions\cite{pomerleau1988alvinn}, with subsequent works improving upon this by incorporating temporal history through various architectures, adopting alternative training objectives, and introducing regularization techniques\cite{jang2022bc,shafiullah2022behavior}. Among recent advances, ACT\cite{zhao2023learning} proposes to predict sequences of actions rather than single-step outputs, reducing the effect of compounding errors and enabling more stable execution of fine-grained manipulation tasks. Diffusion Policy\cite{chi2025diffusion} takes a generative approach by modeling the action distribution as a diffusion process, which effectively captures multimodal behavior and produces smooth, consistent trajectories. Building upon these foundations, subsequent works have achieved notable progress across a wide range of robotic manipulation tasks, further validating the effectiveness of these frameworks\cite{ze20243d,black2025real}.

\section{Methodology}
\label{sec:methods}

\subsection{Hardware and Framework} 
\label{sec:hardware}

\begin{figure*}[t]
    \centering
    \includegraphics[width=1\textwidth]{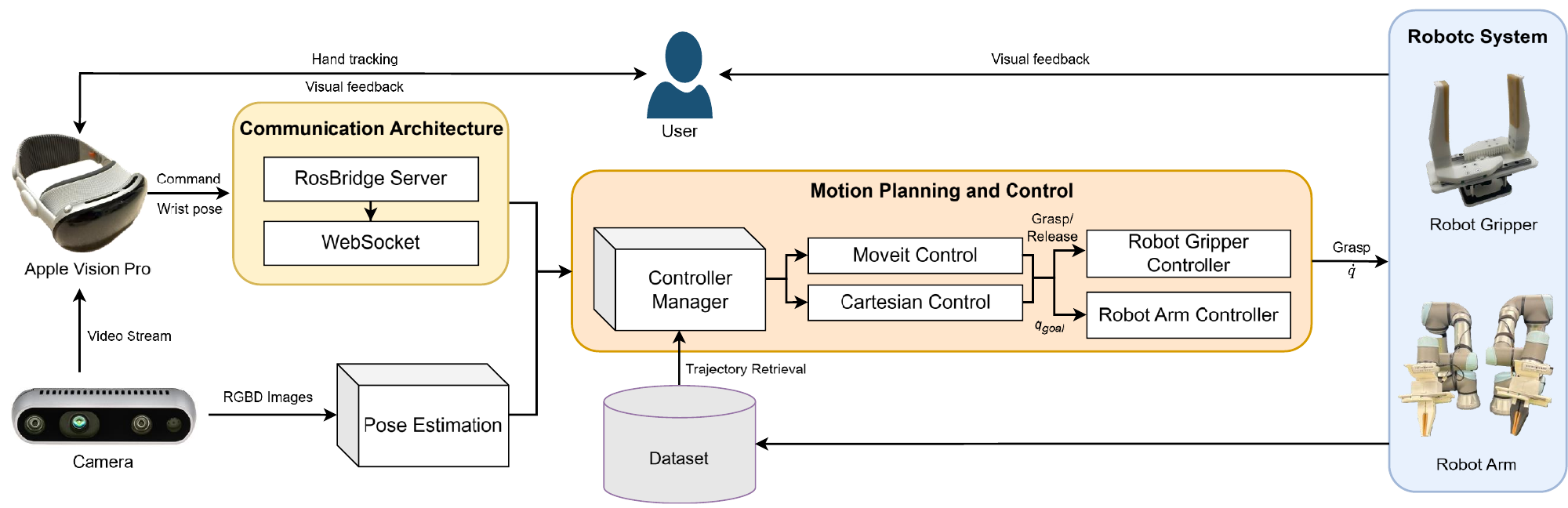}
    \caption{The framework architecture of the MR-based robotic telemanipulation.}
    \label{fig:framework}
\end{figure*}

The hardware platform consists of two 6-DoF UR5e (Universal Robots) serial manipulators, each equipped with a parallel jaw gripper and a wrist-mounted Intel RealSense D435i RGB-D camera. An additional RealSense D435i camera is positioned in front of the workspace to provide a global view of the manipulation scene. This camera is extrinsically calibrated with respect to both robot arms, enabling accurate object pose estimation in the robot coordinate frame. Extrinsic calibration for both the global camera and the wrist-mounted cameras is performed using the easy\_handeye ROS2 package~\cite{tsai1989new}, which implements a standard hand-eye calibration procedure. Both robot arms are managed within a single ROS2 node and share a common system clock, ensuring temporal synchronization between arm states and camera data without requiring additional timestamp alignment at the hardware level. All hardware components are interfaced with an Ubuntu 24.04 workstation running ROS2~\cite{doi:10.1126/scirobotics.abm6074}, which hosts the distributed node architecture responsible for perception, planning, data collection, and robot control.
Fig.\ref{fig:framework} presents an overview of the proposed system framework. The operator interface is built around an AVP mixed reality headset, connected to the ROS2 environment over a local Wi-Fi network. The AVP application is developed in Swift targeting visionOS, and bidirectional communication between the headset and the ROS2 backend is established via the WebSocket-based RosBridge protocol~\cite{crick2016rosbridge}.

\subsection{Bimanual Robotic Telemanipulation}
\label{sec:robot_telem}

In the full telemanipulation modality, human wrist poses and finger gestures are tracked via the AVP using the Tracking Streamer application\cite{park2024avp}. The human wrist poses relative to the AVP are transformed into the robot base frame, mapping the operator's wrist motion to the robot EE in Cartesian space. The EE then follows the target pose by implementing the Cartesian Compliance Controller\cite{FDCC}. Finger gestures are identified by tracking the coordinates of each finger joint. Specifically, the distance between the thumb and index fingertips is measured and mapped to the aperture of the parallel gripper to control its opening and closing. The operator can monitor the real robot directly through the AVP passthrough rendering, providing a real-time, full-color, depth-accurate 3D view of the workspace. To ensure consistent correspondence between the operator and the robot, human-to-robot motion retargeting is implemented so that both robotic arms reliably mirror the operator's left and right arms respectively. The key motivation behind motion retargeting is to maintain behavioral correspondence between the human operator and the robot, since direct kinematic mapping between the human body and the robot structure can otherwise cause significant spatial discrepancies. To achieve this, we record the positions of the human wrists 
$p_{\mathrm{hand}}^0$ and the robot EE $p_{\mathrm{robot}}^0$ at calibration 
to obtain the scale factor $s$ as follows:

\begin{equation}
s = \frac{\|p_{\mathrm{R,robot}}^0 - p_{\mathrm{L,robot}}^0\|}{\|p_{\mathrm{R,hand}}^0 - p_{\mathrm{L,hand}}^0\|}
\end{equation}

The target robot EE position and orientation are then calculated as:

\begin{equation}
p_{\mathrm{robot}}^t = p_{\mathrm{robot}}^0 + s \cdot (p_{\mathrm{hand}}^t - p_{\mathrm{hand}}^0)
\end{equation}

\begin{equation}
R_{\mathrm{robot}}^t = R_{\mathrm{robot}}^0 \cdot A \cdot (R_{\mathrm{hand}}^0)^\top R_{\mathrm{hand}}^t \cdot A^\top
\end{equation}

where $p_{\mathrm{hand}}^t - p_{\mathrm{hand}}^0$ is the hand displacement relative to 
the calibration moment, and $A = (R_{\mathrm{robot}}^0)^\top R_{\mathrm{hand}}^0$ is the 
alignment matrix that transforms the human wrist rotation into the robot frame. 
A low-pass filter is applied to the tracked hand pose to reduce the effect of hand jitter and measurement noise on the resulting robot motion.
This telemanipulation framework is implemented using the Universal Robots ROS2 driver.

\begin{algorithm}[t]
\caption{Trajectory Execution Control Algorithm}
\label{alg:adaptive-velocity}
\KwIn{Increment vector $\vec{\Delta}$, trajectory $T_{\mathrm{current}}$}
\KwOut{Speed $v_{\mathrm{current}}$, trajectory $T_{\mathrm{update}}$}
$v_{\mathrm{normal}} \leftarrow 1$\;
$v_{\mathrm{current}} \leftarrow v_{\mathrm{normal}}$\;
$t_{0} \leftarrow 0$\;
$t_{\mathrm{threshold}} \leftarrow 2$ seconds\;
\While{system active}{
    Acquire $\vec{\Delta}$ from AVP hand tracking\;
    \If{$\|\vec{\Delta}\| > \epsilon$}{
        $v_{\mathrm{current}} \leftarrow 0$\;
        $T_{\mathrm{update}} \leftarrow T_{\mathrm{current}} + \vec{\Delta}$\;
    }
    \Else{
        $t_{0} \leftarrow t_{0} + \Delta t$\;
        \If{$t_{0} \geq t_{\mathrm{threshold}}$}{
            $v_{\mathrm{current}} \leftarrow v_{\mathrm{normal}}$\;
        }
    }
}
\Return $v_{\mathrm{current}}$, $T_{\mathrm{update}}$
\end{algorithm}
\subsection{Semi-Autonomous Control}
\label{sacs}
To leverage the advantages of autonomous precision and human perception, we propose the use of a semi-autonomous telemanipulation framework where task execution is shared between the robot and the operator. In practice, physical contact between the gripper and objects during grasping and placement introduces errors that are difficult to predict, making human intervention at key stages necessary. In order to solve this problem, we implement this shared control approach through two distinct strategies. First, in our Hybrid Telemanipulation (HT) approach, we employ FoundationPose \cite{wen2024foundationpose} so as to estimate the pose of novel objects and generate reliable grasp poses autonomously. Once the object is secured, the operator takes full control via telemanipulation to complete the remainder of the task. Second, we introduce Trajectory Execution Control (TEC). Similar to the HT approach, the robot performs the initial grasp autonomously, but then follows a demonstrated insertion trajectory extracted from an expert demonstration. In this modality, the displacement increments from user input are added onto the current trajectory. When user input is detected, trajectory execution pauses and resumes once no further input is received. The algorithm dynamically regulates the adjustment frequency, providing sufficient reaction time for precise final positioning while maintaining overall efficiency, as detailed in Algorithm~\ref{alg:adaptive-velocity}. 

\begin{figure}[t]
    \centering
    \includegraphics[width=1\columnwidth]{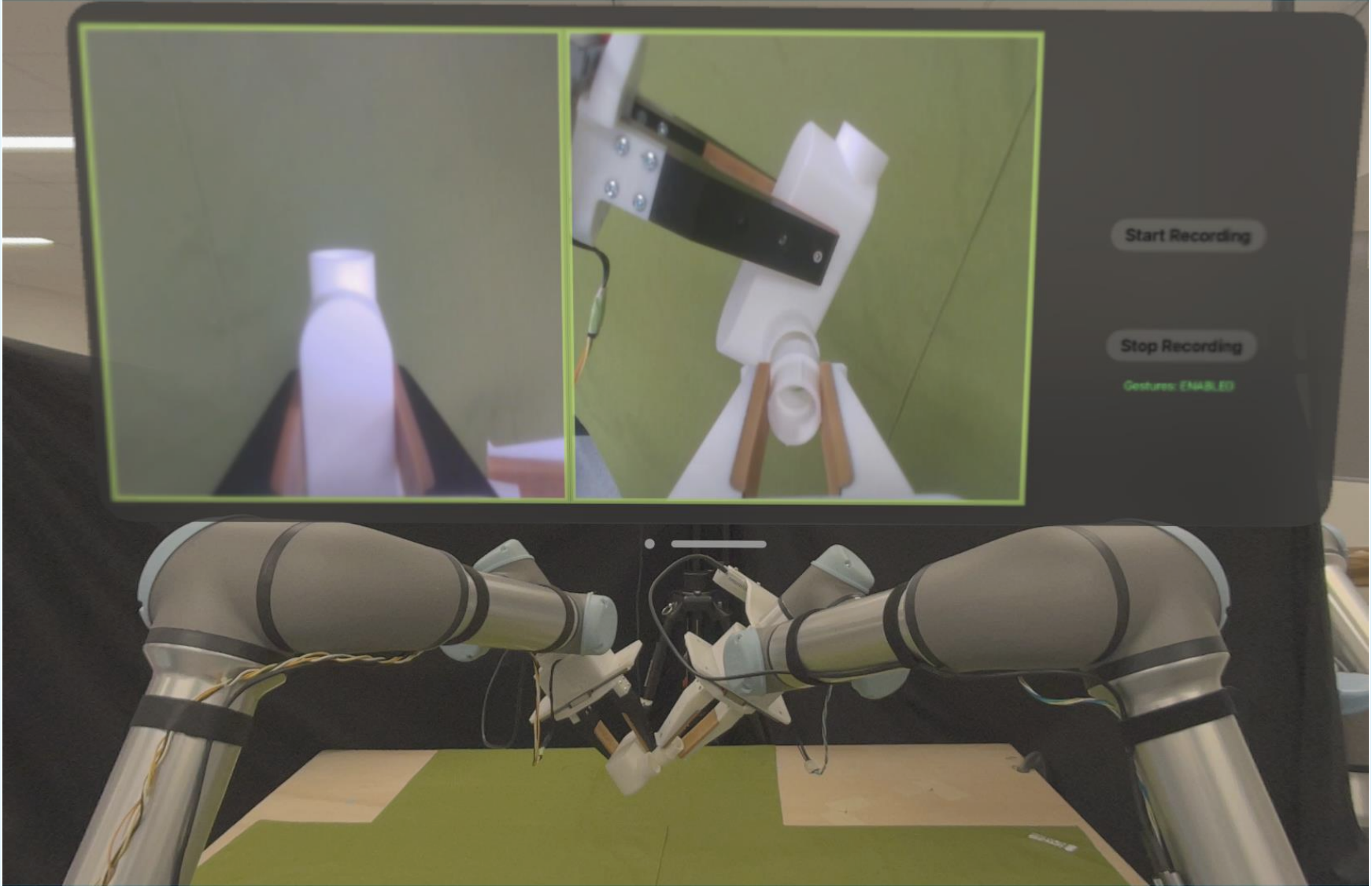}
    \caption{Visualization of the MR environment from the user's perspective.}
    \label{fig:empty}
\end{figure}

Within the MR environment, the user can monitor the robot states via a custom dashboard rendered on the AVP (see Fig. \ref{fig:empty}). This interface features a live video stream from the robot EE camera, alongside two buttons for starting and stopping data collection. The EE can be adjusted via two hand gestures as required: pinch-and-drag motions provide translational control, while bimanual pinching followed by rotational movements controls the EE orientation. User hand movements are captured and their displacement vectors are mapped in real-time from the AVP coordinate frame to robot EE motion commands. 

\begin{figure*}[t]
    \centering
    \includegraphics[width=1\textwidth]{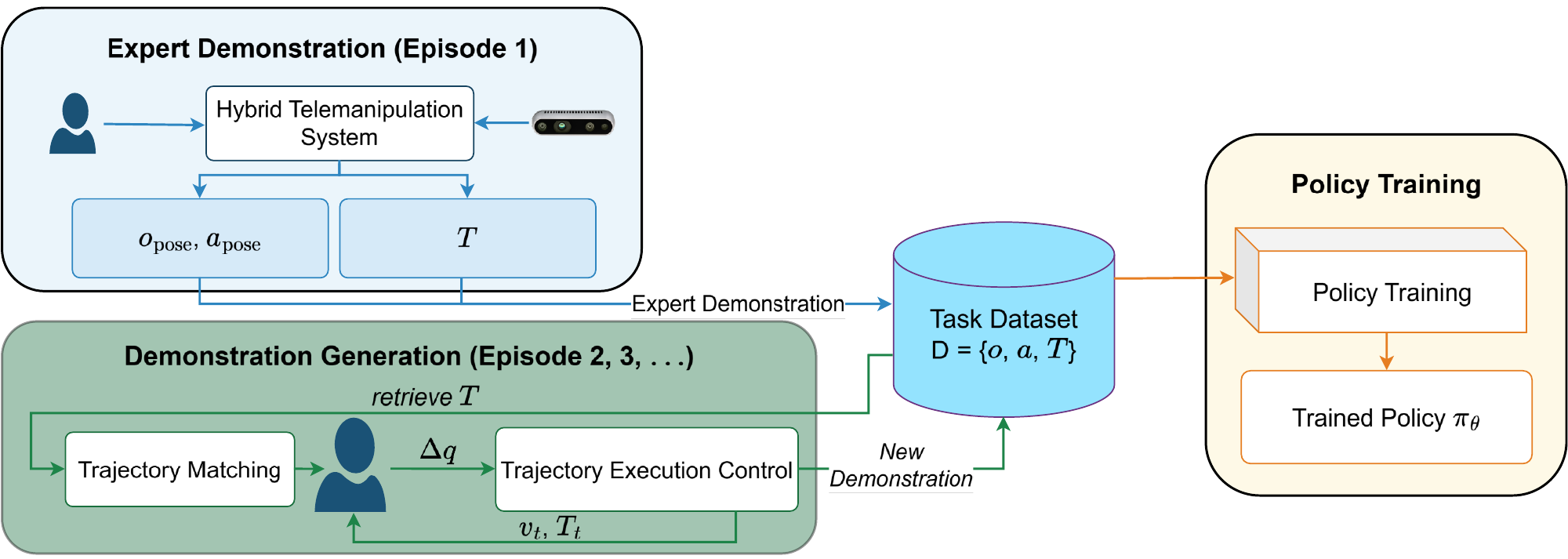}
    \caption{Overview of the imitation learning pipeline integrating with a semi-autonomous telemanipulation system.}
    \label{fig:pipeline}
\end{figure*}

\subsection{Imitation Learning Pipeline}

To enable the execution of fine-grained manipulation tasks, we develop an imitation learning pipeline that integrates our semi-autonomous approach in Fig.\ref{fig:pipeline}. Unlike prior state-of-the-art pipelines that require dozens of expert demonstrations, our approach requires only a single high-quality expert demonstration as a template. During data collection, the robot first performs vision-based path planning as described in Section~\ref{sacs} to autonomously execute the grasping stage, ensuring stable and consistent grasping performance. Once grasping is complete, the operator takes over to telemanipulate the robot arm through the fine-grained manipulation phase. From the second episode onward, the operator can employ the TEC strategy, in which the robot autonomously reproduces the trajectory extracted from the expert demonstration while the operator only needs to make minor adjustments to the robot EE so as to successfully complete the task.

Data collected within each episode are synchronized via the use of appropriate ROS2 message filters for timestamp alignment and post-processed for subsequent training. The observation space consists of RGB images collected from three separate cameras, each with a $480 \times 640$ resolution, and the current robot EE poses of both robotic arms with respect to the robot base frame. The images are processed with ResNet18 backbones so as to extract visual features\cite{he2016deep}. The action space is defined in Cartesian space using the target EE poses. For the dual-arm UR5e platform, each arm contributes 3D position, 6D rotation, and 1 gripper state, giving a 20-dimensional action space. Actions and observations are normalized before training. We implement 6D rotation representation rather than quaternions, as quaternions suffer from double cover, which introduces mathematical discontinuities that can destabilize neural network training. The 6D rotation provides a continuous mapping from Euclidean space to the rotation manifold, ensuring smoother convergence during policy learning\cite{zhou2019continuity}.

\section{Experiments}
\label{sec:userstudy}

\begin{figure*}[t!]
    \centering
    \includegraphics[width=1\textwidth]{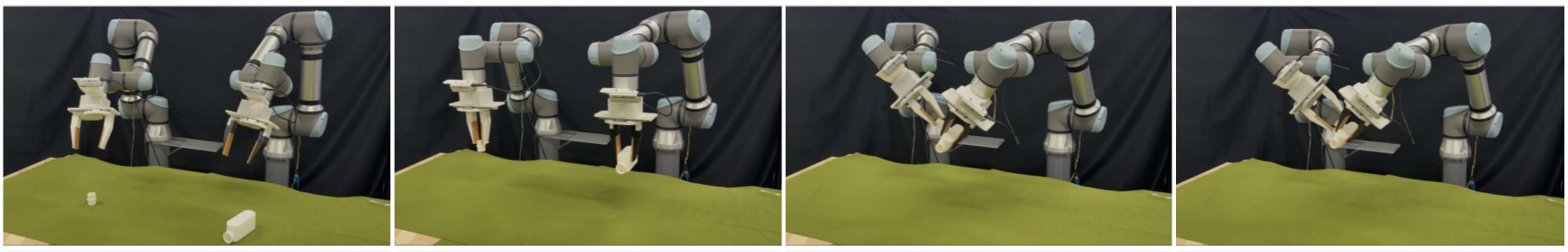}
    \caption{Full demonstration of the execution of the selected bimanual, fine-grained, peg-in-hole task.}
    \label{fig:experiment}
\end{figure*}

\begin{figure}[t]
    \centering
    \includegraphics[width=0.9\columnwidth]{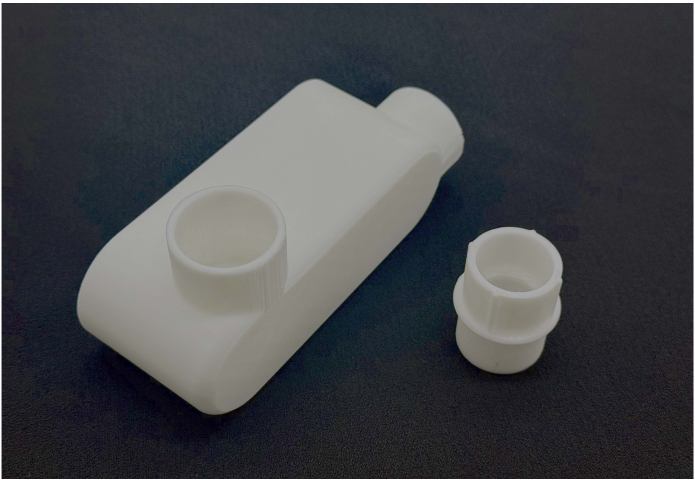}
    \caption{Enclosure and connector used for the execution of the task.}
    \label{fig:peg}
\end{figure}

\subsection{Task and Data Collection}
A bimanual peg-in-hole task - widely regarded as one of the most challenging benchmarks in robotic assembly due to its high precision requirements - is selected for the experimental validation. As shown in Fig. \ref{fig:experiment}, both the enclosure and the connector are placed at random locations on the table within the robot's workspace. The objective is for the robot to grasp both objects and precisely insert the connector into the enclosure with a 1 mm tolerance (Fig. \ref{fig:peg}). The task must be completed without dropping either object, demanding highly coordinated and fine-grained bimanual manipulation. To fully evaluate our system and the proposed TEC strategy, three data collection modalities are compared. The first modality follows Conventional Telemanipulation (CT), consistent with prior works \cite{cheng2024open, iyer2024open, ding2025bunny}, in which the operator fully controls both robot arms and EEs. This mode serves as our experimental baseline. In this setting, the grasping pose varies across episodes, leading to inconsistent initial conditions for the insertion phase. The second modality HT incorporates vision-based path planning for automated grasping, followed by full telemanipulation to complete the insertion. The third modality applies the TEC strategy, in which grasping is performed autonomously and the operator only provides fine adjustments during the insertion phase. For both HT and TEC modalities, only demonstrations after the autonomous grasping stage are collected.

Efficient data collection is a primary limitation for learning-based manipulation, particularly in the execution of complex and long-horizon tasks, as such tasks often lead to failure and operator fatigue. To address this, we evaluate the impact of autonomous grasping on improving demonstration efficiency by comparing HT with the baseline. Furthermore, comparing HT and TEC enables a quantitative assessment of the performance gains derived from human-in-the-loop trajectory correction. For each modality, 30 successful demonstrations were collected at a control frequency of 50 Hz, following prior single-task imitation learning works that trained effective policies with 30--50 demonstrations per task \cite{zhao2023learning,cheng2024open}. To evaluate and compare the performance across modalities, the task completion time and number of failures are recorded.

\subsection{Policy Training and Deployment}

\begin{figure*}[t!]
    \centering
    \includegraphics[width=1\textwidth]{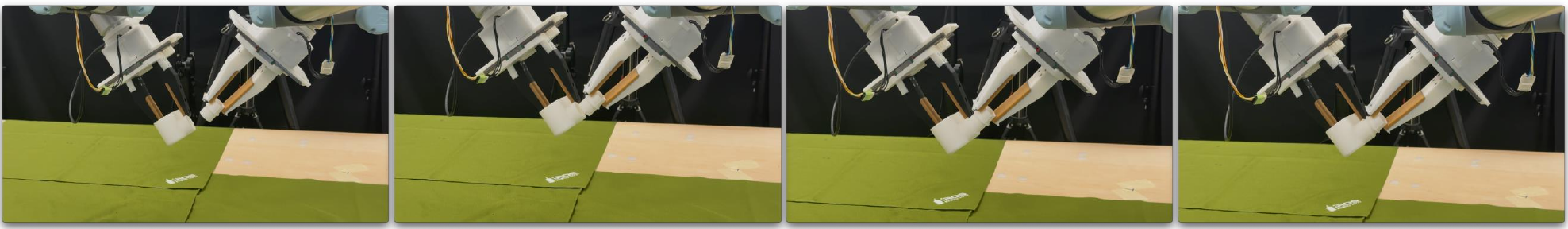}
    \caption{Adjusting the robot end-effector pose via the TEC strategy.}
    \label{fig:task}
    
    \vspace{\floatsep}

    \begin{minipage}[t]{0.49\textwidth}
        \centering
        \includegraphics[width=1\textwidth]{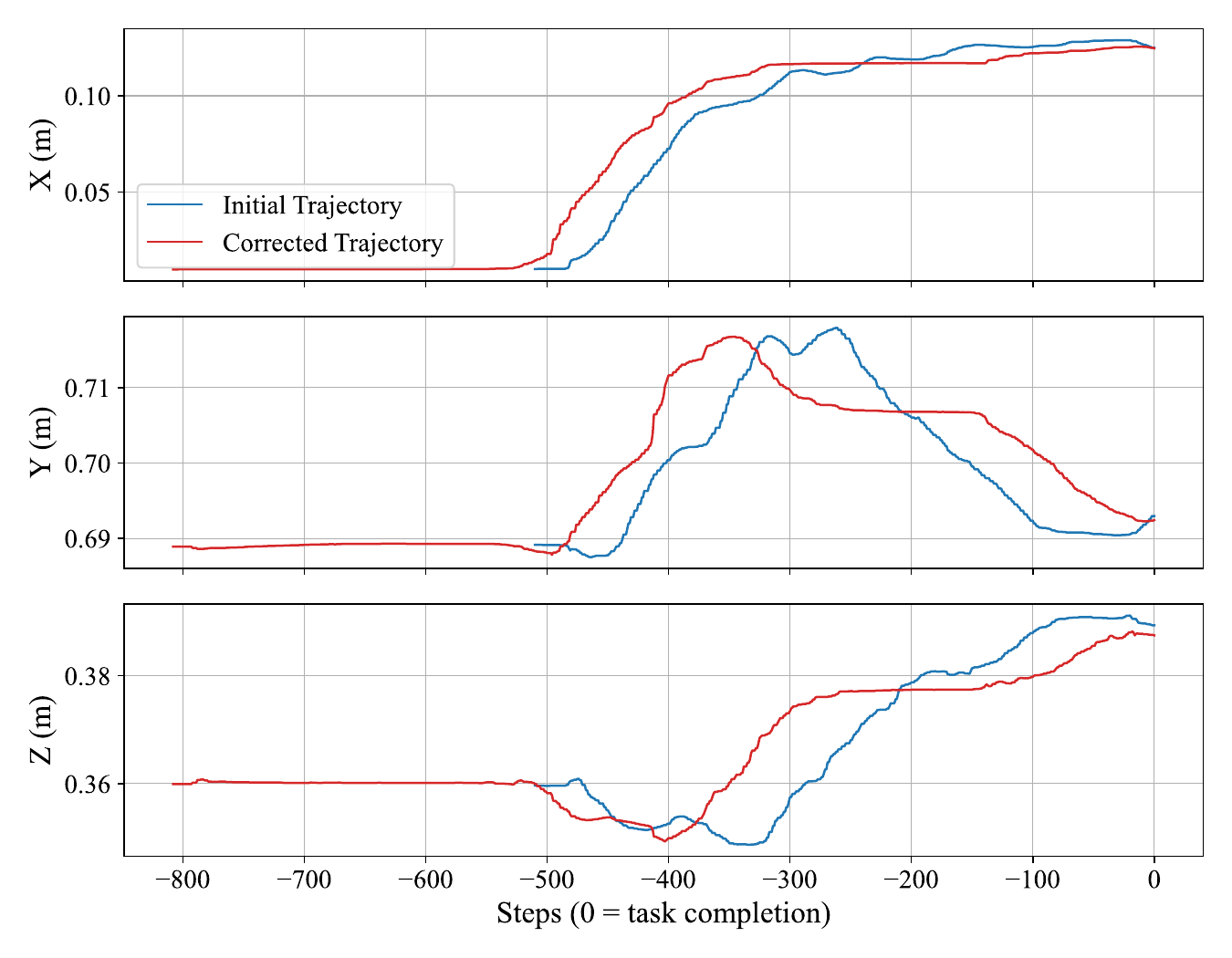}
        \caption{Left arm trajectory: Initial vs. Corrected.}
        \label{fig:trajectory1}
    \end{minipage}
    \hfill
    \begin{minipage}[t]{0.49\textwidth}
        \centering
        \includegraphics[width=1\textwidth]{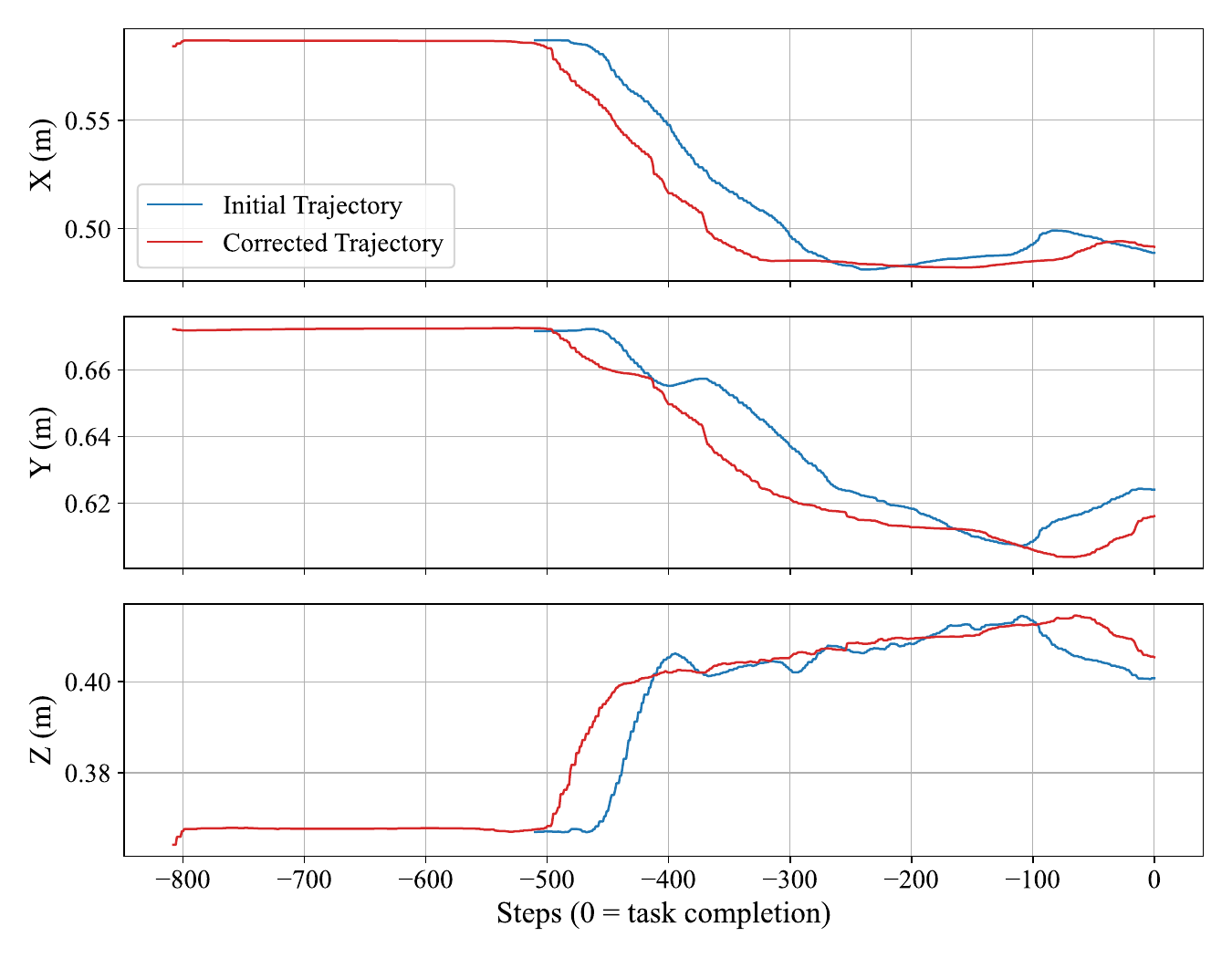}
        \caption{Right arm trajectory: Initial vs. Corrected.}
        \label{fig:trajectory2}
    \end{minipage}
\end{figure*}

To assess the quality of the demonstrations collected by our system, we evaluate its impact on imitation learning against a CT approach. The model has approximately 85M parameters and is trained for around 6 hours on a single 10GB RTX 3080 GPU. We apply identical hyperparameters across all conditions to ensure a fair comparison (Table~\ref{tab:act}). During the evaluation phase, the learned policies for all three modalities are deployed in a fully autonomous mode. Specifically, the deployment for HT and TEC follows the semi-autonomous pipeline described in Section~\ref{sacs}. Once the objects are grasped and secured, the system switches to the trained ACT policy for autonomous assembly instead of telemanipulation. In contrast, for the CT baseline, the learned policy controls the entire process from initial grasping to final assembly without such a transition. To ensure execution stability, temporal ensembling is adopted at an inference frequency of 50 Hz to smooth overlapping action predictions. Each trained policy is evaluated over 30 rollouts under the same experimental conditions as data collection. A trial is considered successful if the connector is fully inserted into the enclosure without dropping. The success rate and execution time are reported as the primary evaluation indicators across all three modalities.

\begin{table}[t]
\centering
\caption{Hyperparameters of ACT}
\label{tab:act}
\begin{tabular}{lc}
\toprule
\textbf{Hyperparameter} & \textbf{Value} \\ 
\midrule
Learning rate         & 1e-5 \\
Batch size            & 8 \\
KL weight             & 10 \\
Feedforward dimension & 3200 \\
Hidden dimension      & 512 \\
Chunk size            & 50 \\
Epochs          & 40000 \\
\bottomrule
\end{tabular}
\end{table}

\section{Results and Discussion}
\label{sec:results}

Fig. \ref{fig:trajectory1} and Fig. \ref{fig:trajectory2} illustrate the EE trajectories of both robot arms, comparing the initial trajectory extracted from the expert demonstration with the corrected trajectory produced by the TEC strategy. The left and right arms correspond to their positions as seen in Fig.\ref{fig:task}. During the connector insertion phase, a single adjustment to one arm is generally sufficient to complete the task. Both trajectories follow a similar path throughout the insertion, with the left arm maintaining its position at the final stage while the right arm shows a slight positional shift, indicating that the TEC strategy applies targeted adjustment where necessary.




\begin{table*}[t]
\centering
\begin{threeparttable}
\caption{Records of Successes, Failures, and Average Execution Time in Data Collection}
\label{tab:collection}
\begin{tabular*}{0.95\textwidth}{@{\extracolsep{\fill}} l ccccc c @{}} 
\toprule
& \multicolumn{2}{c}{\textbf{Failures}} & \textbf{Trials} & \textbf{Overall} & \textbf{Insertion} & \textbf{Avg. Time} \\
\cmidrule(lr){2-3}
& \textbf{Grasp (E / C / B)$^\ast$} & \textbf{Insert} & \textbf{(Success / Total)} & \textbf{Success Rate} & \textbf{Success Rate} & \textbf{(s)} \\
\midrule
CT (Baseline) & \textbf{0 / 1 / 0}  & 7 & 30/38 & 0.78 & 0.81 & {\boldmath $35.5\pm7.21$} \\
Hybrid Telemanipulation      & 1 / 2 / 3  & 3 & 30/39 & 0.77 & 0.91 & $36.5 \pm 5.14$ \\
Trajectory Execution Control      & 0 / 2 / 3 & \textbf{0} & \textbf{30/35} & \textbf{0.86} & \textbf{1.00} & $41.3\pm4.81$ \\
\bottomrule
\end{tabular*}
\begin{tablenotes}
    \small
    \item $^\ast$ E: Enclosure only; C: Connector only; B: Both objects.
\end{tablenotes}
\end{threeparttable}
\end{table*}

Table~\ref{tab:collection} summarizes the success rates, failure counts, and average execution times across the different data collection modalities. Failures are categorized according to the specific stage of the task. Specifically, Insertion failures refer to cases where the objects establish initial grasp, yet the robot fails to complete the final insertion. Since grasping in HT and TEC is autonomous, an insertion success rate, computed over trials that reached the insertion stage, is also reported to isolate the contribution of the telemanipulation system.
The CT approach makes fewer mistakes during the grasping phase compared with the HT and TEC approaches. This is because the user maintains direct visual observation of the workspace, allowing for real-time adjustments to correct any positioning deviations between the gripper and the object. In contrast, the HT and TEC approaches demonstrate similar performance in this stage, as they both rely on pose estimation to localize the objects. Any errors during this process typically result from inaccuracies in the pose estimation and autonomous path planning failure.
However, TEC demonstrates a clear advantage during the insertion phase. While CT and HT require users to manually finalize the insertion, this process is challenging due to the requirement for millimeter level precision and the lack of local visual feedback. HT experiences fewer failures in this stage because the autonomous grasping provides a solid hold, reducing the difficulty for the user. TEC further addresses this by allowing the user to concentrate exclusively on adjusting a single arm. This significantly improves insertion stability, and no failures were observed during our trials.
Furthermore, CT and HT are more time-efficient than TEC as human operators still control most of the execution. As operators gain proficiency through repeated trials, their reaction times and execution speed improve significantly. In contrast, the TEC strategy requires the longest execution time due to its adaptive speed control mechanism. Whenever the user intervenes for fine-tuning, the robot is paused to enable precise alignment. Therefore, more frequent manual corrections introduce more pauses, which extends the task duration. However, the highest overall success rate highlights its robustness. This is further reflected in the insertion success rate, which reaches 1.00 for TEC, 0.91 for HT, and 0.81 for CT, isolating the contribution of the telemanipulation system from autonomous grasping. Although individual demonstrations take longer, TEC enables operators to obtain a sufficient number of successful samples in fewer trials.

\begin{table*}[t]
\centering
\begin{threeparttable}
\caption{Performance of the Learned Policies during Inference: Successes, Failures, and Execution Time}
\label{tab:inference}
\begin{tabular*}{0.95\textwidth}{@{\extracolsep{\fill}} l ccccc c @{}} 
\toprule
& \multicolumn{2}{c}{\textbf{Failures}} & \textbf{Trials} & \textbf{Overall} & \textbf{Insertion} & \textbf{Avg. Time} \\
\cmidrule(lr){2-3}
& \textbf{Grasp (E / C / B)$^\ast$}  & \textbf{Insert} & \textbf{(Success / Total)} & \textbf{Success Rate} & \textbf{Success Rate} & \textbf{(s)} \\
\midrule
CT (Baseline) & 4 / 4 / 5 & 16 & 1/30 & 0.03 & 0.06 & $64.13\pm 12.18$ \\
Hybrid Telemanipulation      & \textbf{0 / 3 / 0} & 7 & 20/30 & 0.67 & 0.74 & {\boldmath$52.7\pm7.17$} \\
Trajectory Execution Control      & 0 / 2 / 2 & \textbf{3} & \textbf{23/30} & \textbf{0.77} & \textbf{0.88} & $53.8\pm 6.39$ \\
\bottomrule
\end{tabular*}
\begin{tablenotes}
    \small
    \item $^\ast$ E: Enclosure only; C: Connector only; B: Both objects.
\end{tablenotes}
\end{threeparttable}
\end{table*}

Table~\ref{tab:inference} illustrates the inference performance of the ACT policies across the three modalities. CT results in the highest failure rate in real-world deployment. One potential reason is the limited number of demonstrations. Furthermore, because objects are placed randomly on the table, they are sometimes only captured by the main camera, providing fewer visual features for the policy. The low CT overall success rate reveals that random grasping poses force the policy to spend more time to grasp the object and offset insertion errors. In some cases, these compounding errors significantly slow down execution. Due to the limited number of successful trials, we define execution time as the duration until the policy ceases to output updates. The longest inference time recorded for CT further supports this finding. In contrast, HT and TEC achieve comparable performance in both overall success rate and execution time, with overall success rates of 0.67 and 0.77 respectively. When grasping failures are excluded, the insertion success rate widens this gap further, with TEC at 0.88, HT at 0.74, and CT at 0.06. TEC yields a higher number of successful trials, likely because the demonstrations include a higher density of steps during the final insertion stage. This provides the policy with finer-grained feature representations, allowing it to more effectively capture the subtle adjustments required for a successful task completion. Furthermore, all three modalities exhibit smooth and continuous trajectories, without observable jitter or jerkiness during the task execution. Overall, HT and TEC achieve improved efficiency and robustness in this high-accuracy, fine-grained bimanual peg-in-hole task execution, where semi-autonomous telemanipulation plays an essential role.

\section{Conclusions and Limitations}
\label{sec:conclusion}

In this paper, we introduced a scalable, intuitive, MR-based semi-autonomous robotic telemanipulation system. By leveraging an AVP headset, the framework integrates vision-based autonomous grasping with human-in-the-loop precision control. Additionally, the system is evaluated through extensive demonstration collection to exhibit its stability and scalability. A real-world deployment using a learned policy for a fine-grained bimanual peg-in-hole task highlights that our system can be effectively utilized by imitation learning algorithms. This approach provides a pathway to scale data collection by minimizing the manual effort required for on-robot demonstrations. By integrating autonomous operation with human perception during the execution of complex manipulation tasks, it offers a practical balance for industrial applications where both efficiency and precision are required.

Despite these advantages, the system has several limitations. Firstly, for contact-rich tasks such as the peg-in-hole insertion phase, where contact forces guide final alignment, the lack of force feedback limits the policy's ability to detect and respond to contact states. Incorporating force/torque sensors during data collection and integrating force information into the imitation learning pipeline could provide richer contact feedback, enabling more adaptive and precise control during deployment. Secondly, while the system is evaluated under varied object orientations, the demonstrations are collected exclusively with objects placed in a facing-up configuration. The generalization to flipped objects or switched object positions remains untested, which may limit the system's robustness in less controlled real-world scenarios. Finally, the number of demonstrations used for training may be insufficient, which is particularly evident in the CT modality. Collecting a larger and more diverse dataset would likely improve the success rate of the learned policy.





\bibliographystyle{IEEEtran}
\bibliography{reference}

\begin{thebibliography}{10}
\providecommand{\url}[1]{#1}
\csname url@samestyle\endcsname
\providecommand{\newblock}{\relax}
\providecommand{\bibinfo}[2]{#2}
\providecommand{\BIBentrySTDinterwordspacing}{\spaceskip=0pt\relax}
\providecommand{\BIBentryALTinterwordstretchfactor}{4}
\providecommand{\BIBentryALTinterwordspacing}{\spaceskip=\fontdimen2\font plus
\BIBentryALTinterwordstretchfactor\fontdimen3\font minus \fontdimen4\font\relax}
\providecommand{\BIBforeignlanguage}[2]{{%
\expandafter\ifx\csname l@#1\endcsname\relax
\typeout{** WARNING: IEEEtran.bst: No hyphenation pattern has been}%
\typeout{** loaded for the language `#1'. Using the pattern for}%
\typeout{** the default language instead.}%
\else
\language=\csname l@#1\endcsname
\fi
#2}}
\providecommand{\BIBdecl}{\relax}
\BIBdecl

\bibitem{brohan2022rt}
A.~Brohan, N.~Brown, J.~Carbajal, Y.~Chebotar, J.~Dabis, C.~Finn, K.~Gopalakrishnan, K.~Hausman, A.~Herzog, J.~Hsu \emph{et~al.}, ``Rt-1: Robotics transformer for real-world control at scale,'' \emph{arXiv preprint arXiv:2212.06817}, 2022.

\bibitem{iyer2024open}
A.~Iyer, Z.~Peng, Y.~Dai, I.~Guzey, S.~Haldar, S.~Chintala, and L.~Pinto, ``Open teach: A versatile teleoperation system for robotic manipulation,'' \emph{arXiv preprint arXiv:2403.07870}, 2024.

\bibitem{lin2025learning}
T.~Lin, Y.~Zhang, Q.~Li, H.~Qi, B.~Yi, S.~Levine, and J.~Malik, ``Learning visuotactile skills with two multifingered hands,'' in \emph{2025 IEEE International Conference on Robotics and Automation (ICRA)}.\hskip 1em plus 0.5em minus 0.4em\relax IEEE, 2025, pp. 5637--5643.

\bibitem{qin2023anyteleop}
Y.~Qin, W.~Yang, B.~Huang, K.~Van~Wyk, H.~Su, X.~Wang, Y.-W. Chao, and D.~Fox, ``Anyteleop: A general vision-based dexterous robot arm-hand teleoperation system,'' \emph{arXiv preprint arXiv:2307.04577}, 2023.

\bibitem{caeiro2021systematic}
M.~Caeiro-Rodr{\'\i}guez, I.~Otero-Gonz{\'a}lez, F.~A. Mikic-Fonte, and M.~Llamas-Nistal, ``A systematic review of commercial smart gloves: Current status and applications,'' \emph{Sensors}, vol.~21, no.~8, p. 2667, 2021.

\bibitem{hetrick2020comparing}
R.~Hetrick, N.~Amerson, B.~Kim, E.~Rosen, E.~J. de~Visser, and E.~Phillips, ``Comparing virtual reality interfaces for the teleoperation of robots,'' in \emph{2020 Systems and Information Engineering Design Symposium (SIEDS)}.\hskip 1em plus 0.5em minus 0.4em\relax IEEE, 2020, pp. 1--7.

\bibitem{park2021study}
S.-J. Park, ``A study on sensor-based upper full-body motion tracking on hololens,'' \emph{Journal of The Korea Society of Computer and Information (JKSCI)}, vol.~26, no.~4, pp. 39--46, 2021.

\bibitem{ding2025bunny}
R.~Ding, Y.~Qin, J.~Zhu, C.~Jia, S.~Yang, R.~Yang, X.~Qi, and X.~Wang, ``Bunny-visionpro: Real-time bimanual dexterous teleoperation for imitation learning,'' in \emph{2025 IEEE/RSJ International Conference on Intelligent Robots and Systems (IROS)}.\hskip 1em plus 0.5em minus 0.4em\relax IEEE, 2025, pp. 12\,248--12\,255.

\bibitem{cheng2024open}
X.~Cheng, J.~Li, S.~Yang, G.~Yang, and X.~Wang, ``Open-television: Teleoperation with immersive active visual feedback,'' \emph{arXiv preprint arXiv:2407.01512}, 2024.

\bibitem{alharthi2024towards}
A.~S. Alharthi, O.~Tokatli, E.~Lopez, and G.~Herrmann, ``Towards semi-autonomous robotic arm manipulation operator intention detection from force data,'' \emph{IEEE Access}, 2024.

\bibitem{min2025semi}
D.~Min, H.~Yoon, and D.~Lee, ``A semi-autonomous telemanipulation order-picking control based on estimating operator intent for box-stacking storage environments,'' \emph{Sensors}, vol.~25, no.~4, p. 1217, 2025.

\bibitem{duan2024semi}
S.~Duan, F.~De~Pace, F.~P. Sanches, H.~Jiang, and M.~Liarokapis, ``Semi-autonomous, virtual reality based robotic telemanipulation for the execution of peg-in-hole assembly tasks,'' in \emph{2024 IEEE-RAS 23rd International Conference on Humanoid Robots (Humanoids)}.\hskip 1em plus 0.5em minus 0.4em\relax IEEE, 2024, pp. 351--358.

\bibitem{duan2026comparing}
S.~Duan, F.~De~Pace, and M.~Liarokapis, ``Comparing semi-autonomous strategies for virtual reality based remote robotic telemanipulation: On peg-in-hole tasks,'' \emph{IEEE Robotics and Automation Letters}, vol.~11, no.~3, pp. 2562--2569, 2026.

\bibitem{zhang2018deep}
T.~Zhang, Z.~McCarthy, O.~Jow, D.~Lee, X.~Chen, K.~Goldberg, and P.~Abbeel, ``Deep imitation learning for complex manipulation tasks from virtual reality teleoperation,'' in \emph{2018 IEEE International Conference on Robotics and Automation (ICRA)}.\hskip 1em plus 0.5em minus 0.4em\relax IEEE, 2018, pp. 5628--5635.

\bibitem{delpreto2020helping}
J.~DelPreto, J.~I. Lipton, L.~Sanneman, A.~J. Fay, C.~Fourie, C.~Choi, and D.~Rus, ``Helping robots learn: a human-robot master-apprentice model using demonstrations via virtual reality teleoperation,'' in \emph{2020 IEEE International Conference on Robotics and Automation (ICRA)}.\hskip 1em plus 0.5em minus 0.4em\relax IEEE, 2020, pp. 10\,226--10\,233.

\bibitem{pan2021augmented}
Y.~Pan, C.~Chen, D.~Li, Z.~Zhao, and J.~Hong, ``Augmented reality-based robot teleoperation system using rgb-d imaging and attitude teaching device,'' \emph{Robotics and Computer-Integrated Manufacturing}, vol.~71, p. 102167, 2021.

\bibitem{penco2024mixed}
L.~Penco, K.~Momose, S.~McCrory, D.~Anderson, N.~Kitchel, D.~Calvert, and R.~J. Griffin, ``Mixed reality teleoperation assistance for direct control of humanoids,'' \emph{IEEE Robotics and Automation Letters}, vol.~9, no.~2, pp. 1937--1944, 2024.

\bibitem{duan2026mr}
S.~Duan, F.~De~Pace, Z.~Wang, and M.~Liarokapis, ``Mixed reality-based, immersive, semi-autonomous robotic telemanipulation for the execution of peg-in-hole tasks,'' in \emph{2026 IEEE International Conference on Robotics and Automation (ICRA)}.\hskip 1em plus 0.5em minus 0.4em\relax IEEE, 2026.

\bibitem{li2025comparative}
L.~Li, D.~Crosby, M.~Shuttleworth, O.~F. Argin, A.~S. Chen, G.~Herrmann, R.~Kay, and A.~Weightman, ``A comparative analysis and scoping review of soft--rigid and industrial parallel rigid grippers,'' \emph{Advanced Intelligent Systems}, vol.~7, no.~5, p. 2400503, 2025.

\bibitem{khazatsky2024droid}
A.~Khazatsky, K.~Pertsch, S.~Nair, A.~Balakrishna, S.~Dasari, S.~Karamcheti, S.~Nasiriany, M.~K. Srirama, L.~Y. Chen, K.~Ellis \emph{et~al.}, ``Droid: A large-scale in-the-wild robot manipulation dataset,'' \emph{arXiv preprint arXiv:2403.12945}, 2024.

\bibitem{zitkovich2023rt}
B.~Zitkovich, T.~Yu, S.~Xu, P.~Xu, T.~Xiao, F.~Xia, J.~Wu, P.~Wohlhart, S.~Welker, A.~Wahid \emph{et~al.}, ``Rt-2: Vision-language-action models transfer web knowledge to robotic control,'' in \emph{Conference on Robot Learning}.\hskip 1em plus 0.5em minus 0.4em\relax PMLR, 2023, pp. 2165--2183.

\bibitem{siekmann2021blind}
J.~Siekmann, K.~Green, J.~Warila, A.~Fern, and J.~Hurst, ``Blind bipedal stair traversal via sim-to-real reinforcement learning,'' \emph{arXiv preprint arXiv:2105.08328}, 2021.

\bibitem{zhao2023learning}
T.~Z. Zhao, V.~Kumar, S.~Levine, and C.~Finn, ``Learning fine-grained bimanual manipulation with low-cost hardware,'' \emph{arXiv preprint arXiv:2304.13705}, 2023.

\bibitem{fang2023low}
H.~Fang, H.-S. Fang, Y.~Wang, J.~Ren, J.~Chen, R.~Zhang, W.~Wang, and C.~Lu, ``Low-cost exoskeletons for learning whole-arm manipulation in the wild,'' in \emph{Towards Generalist Robots: Learning Paradigms for Scalable Skill Acquisition@ CoRL2023}, 2023.

\bibitem{fu2024mobile}
Z.~Fu, T.~Z. Zhao, and C.~Finn, ``Mobile aloha: Learning bimanual mobile manipulation using low-cost whole-body teleoperation,'' in \emph{8th Annual Conference on Robot Learning}, 2024.

\bibitem{wu2024gello}
P.~Wu, Y.~Shentu, Z.~Yi, X.~Lin, and P.~Abbeel, ``Gello: A general, low-cost, and intuitive teleoperation framework for robot manipulators,'' in \emph{2024 IEEE/RSJ International Conference on Intelligent Robots and Systems (IROS)}.\hskip 1em plus 0.5em minus 0.4em\relax IEEE, 2024, pp. 12\,156--12\,163.

\bibitem{pomerleau1988alvinn}
D.~A. Pomerleau, ``Alvinn: An autonomous land vehicle in a neural network,'' \emph{Advances in neural information processing systems}, vol.~1, 1988.

\bibitem{jang2022bc}
E.~Jang, A.~Irpan, M.~Khansari, D.~Kappler, F.~Ebert, C.~Lynch, S.~Levine, and C.~Finn, ``Bc-z: Zero-shot task generalization with robotic imitation learning,'' in \emph{conference on Robot Learning}.\hskip 1em plus 0.5em minus 0.4em\relax PMLR, 2022, pp. 991--1002.

\bibitem{shafiullah2022behavior}
N.~M. Shafiullah, Z.~Cui, A.~A. Altanzaya, and L.~Pinto, ``Behavior transformers: Cloning $ k $ modes with one stone,'' \emph{Advances in neural information processing systems}, vol.~35, pp. 22\,955--22\,968, 2022.

\bibitem{chi2025diffusion}
C.~Chi, Z.~Xu, S.~Feng, E.~Cousineau, Y.~Du, B.~Burchfiel, R.~Tedrake, and S.~Song, ``Diffusion policy: Visuomotor policy learning via action diffusion,'' \emph{The International Journal of Robotics Research}, vol.~44, no. 10-11, pp. 1684--1704, 2025.

\bibitem{ze20243d}
Y.~Ze, G.~Zhang, K.~Zhang, C.~Hu, M.~Wang, and H.~Xu, ``3d diffusion policy: Generalizable visuomotor policy learning via simple 3d representations,'' \emph{arXiv preprint arXiv:2403.03954}, 2024.

\bibitem{black2025real}
K.~Black, M.~Y. Galliker, and S.~Levine, ``Real-time execution of action chunking flow policies,'' \emph{arXiv preprint arXiv:2506.07339}, 2025.

\bibitem{tsai1989new}
R.~Y. Tsai, R.~K. Lenz \emph{et~al.}, ``A new technique for fully autonomous and efficient 3 d robotics hand/eye calibration,'' \emph{IEEE Transactions on robotics and automation}, vol.~5, no.~3, pp. 345--358, 1989.

\bibitem{doi:10.1126/scirobotics.abm6074}
\BIBentryALTinterwordspacing
S.~Macenski, T.~Foote, B.~Gerkey, C.~Lalancette, and W.~Woodall, ``Robot operating system 2: Design, architecture, and uses in the wild,'' \emph{Science Robotics}, vol.~7, no.~66, p. eabm6074, 2022. [Online]. Available: \url{https://www.science.org/doi/abs/10.1126/scirobotics.abm6074}
\BIBentrySTDinterwordspacing

\bibitem{crick2016rosbridge}
C.~Crick, G.~Jay, S.~Osentoski, B.~Pitzer, and O.~C. Jenkins, ``Rosbridge: Ros for non-ros users,'' in \emph{Robotics research: The 15th international symposium ISRR}.\hskip 1em plus 0.5em minus 0.4em\relax Springer, 2016, pp. 493--504.

\bibitem{park2024avp}
\BIBentryALTinterwordspacing
Y.~Park and P.~Agrawal, ``Using apple vision pro to train and control robots,'' 2024. [Online]. Available: \url{https://github.com/Improbable-AI/VisionProTeleop}
\BIBentrySTDinterwordspacing

\bibitem{FDCC}
S.~Scherzinger, A.~Roennau, and R.~Dillmann, ``Forward dynamics compliance control (fdcc): A new approach to cartesian compliance for robotic manipulators,'' in \emph{IEEE/RSJ International Conference on Intelligent Robots and Systems (IROS)}, 2017, pp. 4568--4575.

\bibitem{wen2024foundationpose}
B.~Wen, W.~Yang, J.~Kautz, and S.~Birchfield, ``Foundationpose: Unified 6d pose estimation and tracking of novel objects,'' in \emph{Proceedings of the IEEE/CVF Conference on Computer Vision and Pattern Recognition}, 2024, pp. 17\,868--17\,879.

\bibitem{he2016deep}
K.~He, X.~Zhang, S.~Ren, and J.~Sun, ``Deep residual learning for image recognition,'' in \emph{Proceedings of the IEEE conference on computer vision and pattern recognition}, 2016, pp. 770--778.

\bibitem{zhou2019continuity}
Y.~Zhou, C.~Barnes, J.~Lu, J.~Yang, and H.~Li, ``On the continuity of rotation representations in neural networks,'' in \emph{Proceedings of the IEEE/CVF conference on computer vision and pattern recognition}, 2019, pp. 5745--5753.

\end{thebibliography}

\end{document}